# EFormer: Temporally Aligned Local Correction for Continuous sEMG-Based Hand Pose Tracking

**JiaCheng Ge, SiYu Zhang**

**Abstract**

Surface electromyography (sEMG) provides a wearable, camera-free signal for continuous hand-motion inference. Mapping muscle activity to joint kinematics remains challenging because the recorded waveforms are indirect measurements, their relationship with motion changes over time, and individual anatomy and sensor placement alter the signal distribution. This paper presents EFormer, a residual feature-correction network built on a frozen tracking backbone. EFormer combines a high-rate event branch, temporally aligned local cross-attention, two causal rotary position embedding (RoPE) temporal layers, and a bounded, dynamically gated residual.

EFormer receives 16-channel sEMG sampled at 2 kHz and fuses a 64-channel tracking representation at 25 Hz with a 128-channel event representation at 200 Hz. Cross-attention uses a nominal delay of 100 ms, a 300 ms history parameter, and a 50 ms tolerance; its causal mask restricts each query to events occurring 50–400 ms earlier. The correction scale is 0.15. The evaluated continuation-training configuration contains 585,376 trainable parameters and 5,974,508 frozen parameters.

On the test set, EFormer achieves an MAE of 0.1546634 rad, an RMSE of 0.24063 rad, and an $R^2$ of 0.74801, compared with 0.1745326 rad, 0.2715448 rad, and 0.6791103 for the official tracking baseline. EFormer reduces MAE by 11.38% relative to the baseline. The results show that temporally aligned event-feature correction can reduce continuous hand-pose tracking error.

## 1. Introduction

The hand is a high-bandwidth input device for manipulation, communication, rehabilitation, and immersive computing. Wrist-worn sEMG sensors provide a compact means of observing muscle activity without cameras or an unobstructed view of the hand. Unlike optical tracking, sEMG remains available when the hand leaves the camera field of view, illumination is poor, or partial occlusion occurs. The sensors, however, do not observe pose directly. They measure electrical activity associated with muscle activation, from which pose must be inferred through relationships that vary over time and across individuals.

Continuous sEMG-to-pose estimation is particularly affected by three factors. First, each channel mixes activity from multiple muscles, while a movement typically involves coordinated activation of several muscles. Second, electromechanical and behavioral delays separate muscle activation from joint motion. Third, anatomy, electrode placement, contact quality, and movement strategy vary across users and sessions. A model may therefore perform well on familiar recordings yet respond slowly or lose accuracy after a user re-dons the sensor. These challenges have motivated large-scale datasets and highlight the need for held-out evaluation by user, movement stage, or both.

The emg2pose benchmark provides an important setting for this problem. It contains recordings from 193 users, 370 hours of data, and 29 movement stages, using 16-channel wrist sEMG sampled at 2 kHz with high-quality pose labels [1]. Its tasks evaluate generalization to unseen users and movement behavior. Earlier work such as NeuroPose combined learned sequence models with hand-structure constraints [5], while recurrent networks have also mapped sEMG to continuous hand pose [6]. The benchmark clarifies the scale and generalization requirements of the task, but temporal ambiguity in the supervision remains.

EFormer begins with a frozen tracking representation and introduces a temporally constrained high-rate event branch for local feature correction. Local cross-attention associates low-rate tracking features with events from a specified historical interval. Two causal RoPE encoder layers further refine the aligned representation, after which a bounded correction and dynamic gate produce the residual. The feature network outputs a corrected 64-channel tracking representation, which a downstream pose decoder maps to 20 hand-joint degrees of freedom.

The main contributions are: (1) EFormer, which combines a frozen tracking representation with high-rate sEMG event features for local residual correction; (2) physically time-constrained local cross-attention, causal

RoPE temporal refinement, and dynamic gating; and (3) an evaluation on continuous hand-pose prediction using MAE, RMSE, and $R^2$ against the official tracking baseline.

## 2. Background and Related Work

### 2.1 Surface Electromyography as a Hand-Pose Signal

sEMG records electrical potentials associated with muscle activity. It is suitable for wearable interaction because signals can be acquired at the wrist or forearm without optically observing the hand. The main challenge in continuous pose estimation is that the measurement is an indirect and spatially mixed representation of motion. Different movement stages may exhibit similar instantaneous activation patterns, while the same intended movement may produce different signals across individuals or sensor placements.

Continuous pose estimation predates current large-scale benchmarks. Quivira et al. translated sEMG into continuous hand poses with recurrent neural networks and represented uncertainty with a Gaussian mixture model [6]. Liu et al. proposed NeuroPose, combining anatomical constraints with recurrent, encoder-decoder, and residual architectures for three-dimensional hand-pose tracking from wearable EMG [5]. Sîmpetru et al. investigated convolutional deep learning for continuous prediction of 14 hand degrees of freedom [7]. These studies established recurring design choices: using temporal context, encoding hand structure when possible, and evaluating continuous trajectories rather than only discrete gesture labels.

The emg2pose work expands both the available data scale and the benchmark scope. It introduces recordings from 193 users, 370 hours of data, 29 movement stages, 16 channels, and a 2 kHz sampling rate [1], and compares several model families and generalization splits. This benchmark provides a unified task setting for continuous hand-pose estimation.

### 2.2 Temporal Annotation Error and Alignment

Temporal alignment is more than a tuning convenience. Movement cues, participant reaction time, sensor processing, and annotation timelines cannot be synchronized perfectly, so offsets may arise between EMG recordings and kinematic labels. Wang et al. studied global and trial-wise EMG-kinematic realignment, reporting trial-dependent misalignment. Offline realignment improved performance, whereas online closed-loop behavior did not show the same improvement [8]. Gains from offline alignment therefore should not automatically be interpreted as evidence of improved closed-loop behavior.

EFormer addresses this issue within the model. It does not move stored labels; instead, it supplies the cross-attention module with a nominal delay and a finite event history. The distinction is important: label realignment changes the temporal correspondence of the training targets, whereas the EFormer attention mask changes which event features can contribute to the representation at a tracking time. The two approaches may be studied together, but their mechanisms and causal interpretations remain distinct.

### 2.3 Sequence Models and Positional Information

Recurrent networks provide explicit temporal state and have been used for EMG decoding [5,6]. Temporal convolutional networks support parallel computation through causal filters and progressively expanding receptive fields; the empirical comparison by Bai et al. showed that generic convolutional sequence models can be strong alternatives to recurrent networks [4]. The emg2pose baseline also uses time-depth separable convolutions, a design studied by Hannun et al. for efficient sequence recognition [9]. Transformer self-attention permits direct interaction between sequence positions [2], but its temporal interpretation depends on positional encoding and masking. RoPE applies position-dependent rotations to queries and keys so that attention scores encode relative position [3].

EFormer combines these ideas under explicit constraints. The event-to-tracking operation is local and limited by a temporal mask; the subsequent temporal encoder is causal and uses rotary positional information. Causality follows from the masking and context settings rather than from RoPE itself. This distinction matters in streaming applications: positional encoding can express temporal relationships, but it cannot prevent access to future samples without an appropriate causal mask.

### 2.4 Adaptation and Calibration

Personalization methods address distribution shifts caused by users and sensor placement. The 2026 REACT preprint proposes a conditioning framework for user-adaptive sEMG hand-pose estimation, using compact user embeddings and feature-wise modulation of a frozen backbone [10]. EFormer instead corrects tracking representations with temporally aligned event features, addressing a different modeling factor.

### 2.5 Positioning of EFormer

EFormer is a residual correction network attached to an existing sEMG tracking representation. Its high-rate event branch and local temporal alignment allow low-rate tracking features to use recent muscle-activity information; a downstream pose decoder maps the resulting features to joint angles. EFormer improves all reported metrics, including MAE, RMSE, and $R^2$, over the official tracking baseline.

## 3. Task Definition and Notation

Let the sampled sEMG sequence be $X = \{x_t\}$, where $x_t \in R^{16}$ and the sampling rate is 2,000 Hz. Let $y_t$ denote the system-level hand-configuration target with 20 degrees of freedom. EFormer receives the sEMG sequence and outputs a corrected 64-channel tracking representation at the baseline temporal resolution; a downstream pose decoder maps this representation to joint angles.

The task is causal sequence estimation: the output at time t may depend on input samples at and before t, but not on future samples. The central alignment question is which past event tokens a tracking query at time t may access. The event representation operates at 200 Hz and the baseline tracking representation at 25 Hz. The nominal delay, history parameter, and tolerance are 100 ms, 300 ms, and 50 ms, respectively. Under this temporal mask, a query can access events from t−400 ms to t−50 ms and receives no future information.

For a test set containing M scalar joint-angle observations, the mean absolute error is defined as:

$$\mathrm{MAE} = (1/M) \times \Sigma_i |\hat{y}_i - y_i|.$$

Mean absolute error (MAE) is the primary metric; root mean square error (RMSE) and the coefficient of determination $R^2$ are also reported. MAE and RMSE are measured in radians, while $R^2$ quantifies the extent to which the predictions explain variation in the target.

Relative error reduction is computed as (baseline MAE − EFormer MAE)/baseline MAE. RMSE and $R^2$ are compared in their respective favorable directions.

## 4. EFormer Architecture

### 4.1 Architecture Overview

EFormer consists of a frozen tracking backbone and a trainable residual-correction branch. The model uses 16 input channels, 64 baseline output-feature channels, a 128-channel event representation, a 128-dimensional attention space, four attention heads, two temporal layers, a correction scale of 0.15, a 100 ms delay, a 300 ms history, a 50 ms tolerance, and a 2 kHz input sampling rate. The forward pass computes baseline and event sequences, followed by local cross-attention, causal temporal refinement, bounded correction, and a gated feature residual.

Figure 1 summarizes this data flow. Arrows indicate tensor paths, and labels report channel counts and nominal rates. The diagram is architectural; the downstream 20-DoF decoder is shown for context and is not part of the feature network.

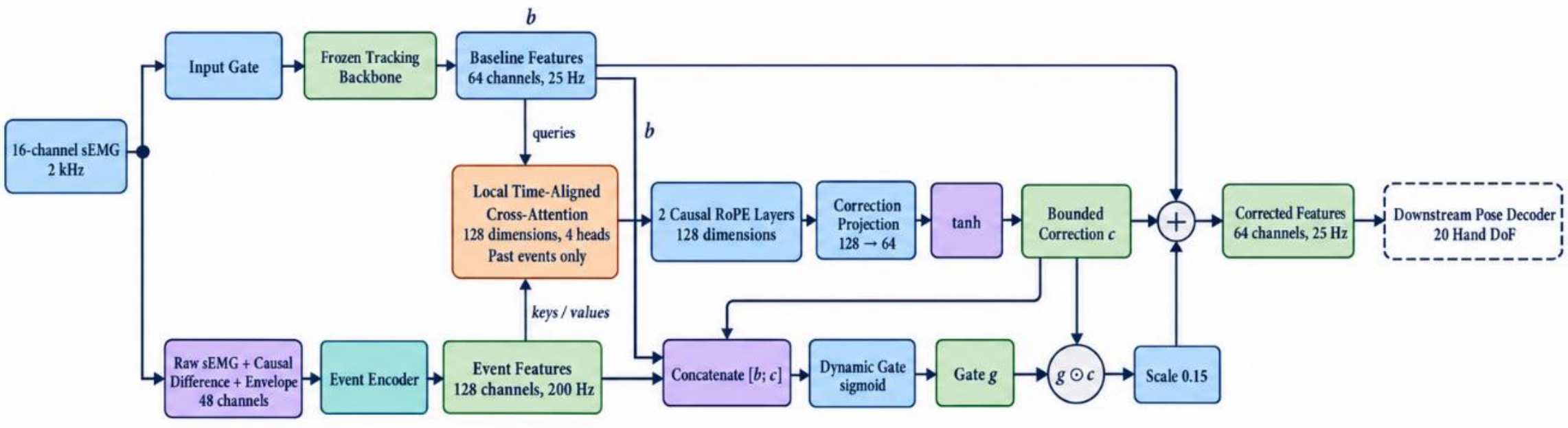

Figure 1. EFormer feature-correction architecture.

The computation can be written as:

$b = B(G(x)),\ e = E(x),\ z = A_local(b, e);$

$h = C_2(C_1(z)),\ c = \tanh(W_c\ h);$

$g = \mathrm{sigmoid}(W_g[b; c]),\ \hat{f} = b + s(g \odot c),\ s = 0.15.$

Here, B denotes the frozen baseline tracking path, G the input gate, E the high-rate event encoder, A_local temporally aligned local cross-attention, $C_1$ and $C_2$ the causal RoPE layers, W_c the correction projection, and W_g the dynamic-gate projection. The final output has the same channel count and temporal resolution as the baseline representation.

### 4.2 Frozen Tracking Path

The frozen tracking path outputs 64-channel features and uses a left context of 1,790 samples with zero right context. At a 2 kHz sampling rate, the zero right context works with the causal masks in the event branch and temporal layers to limit the information available at each time step.

Freezing the baseline has two consequences. First, EFormer uses the existing tracking representation as a stable reference. Second, the trainable correction branch can be optimized without updating all baseline parameters. The evaluated continuation-training configuration contains 585,376 trainable parameters and 5,974,508 frozen parameters.

### 4.3 High-Rate Event Encoding

The event input concatenates raw sEMG, a causal first difference, and a 32-sample absolute-value envelope, producing 48 channels. At a 2 kHz sampling rate, 32 samples correspond to 16 ms. The event encoder maps the input to a 128-channel representation with dropout 0.05 and a stride of 10 input samples, yielding 200 event frames per second.

The first difference provides local change information, while the short-term absolute-value envelope summarizes recent signal magnitude. Together with raw sEMG, these components allow the event encoder to use both amplitude and change trends. Figure 2 shows the event-input channel composition.

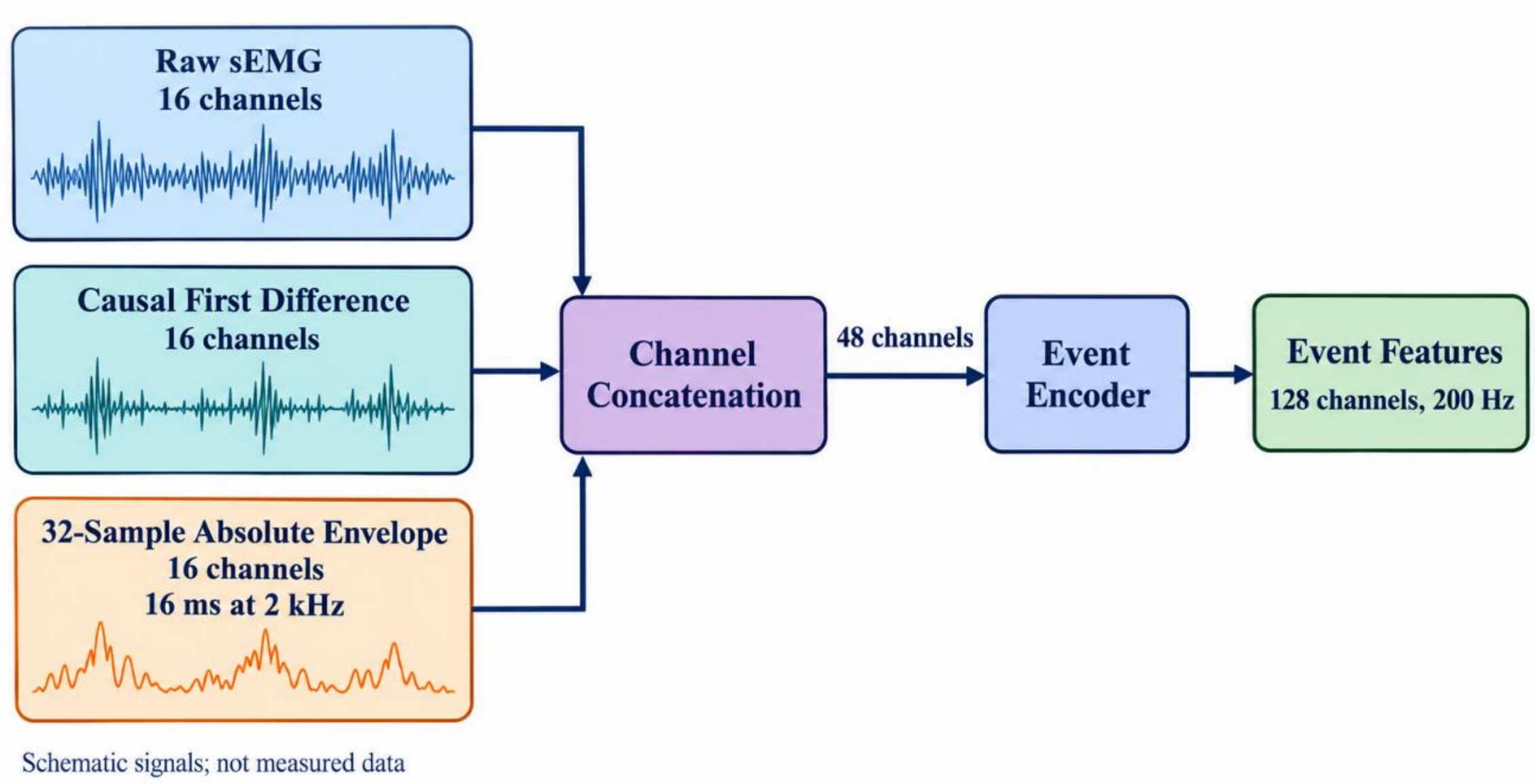

Figure 2. Channel construction from the raw signal, causal difference, and short-term envelope. The diagram does not depict measured data.

### 4.4 Multi-Rate Representations and Local Cross-Attention

With a tracking stride of 80 samples at a 2 kHz input rate, the baseline path produces a 64-channel representation at 25 Hz. The event branch has a stride of 10 and therefore produces a 128-channel representation at 200 Hz. Local cross-attention receives both representations, projects them into 128 dimensions, uses four attention heads, and includes a feed-forward layer of width 256 with dropout 0.10.

At 2 kHz, the 100 ms nominal delay corresponds to 200 samples, the 300 ms history to 600 samples, and the 50 ms tolerance to 100 samples. The local-attention event window spans 50–400 ms before the query, and future events are masked. Figure 3 shows this interval and the rate difference between the event and tracking clocks.

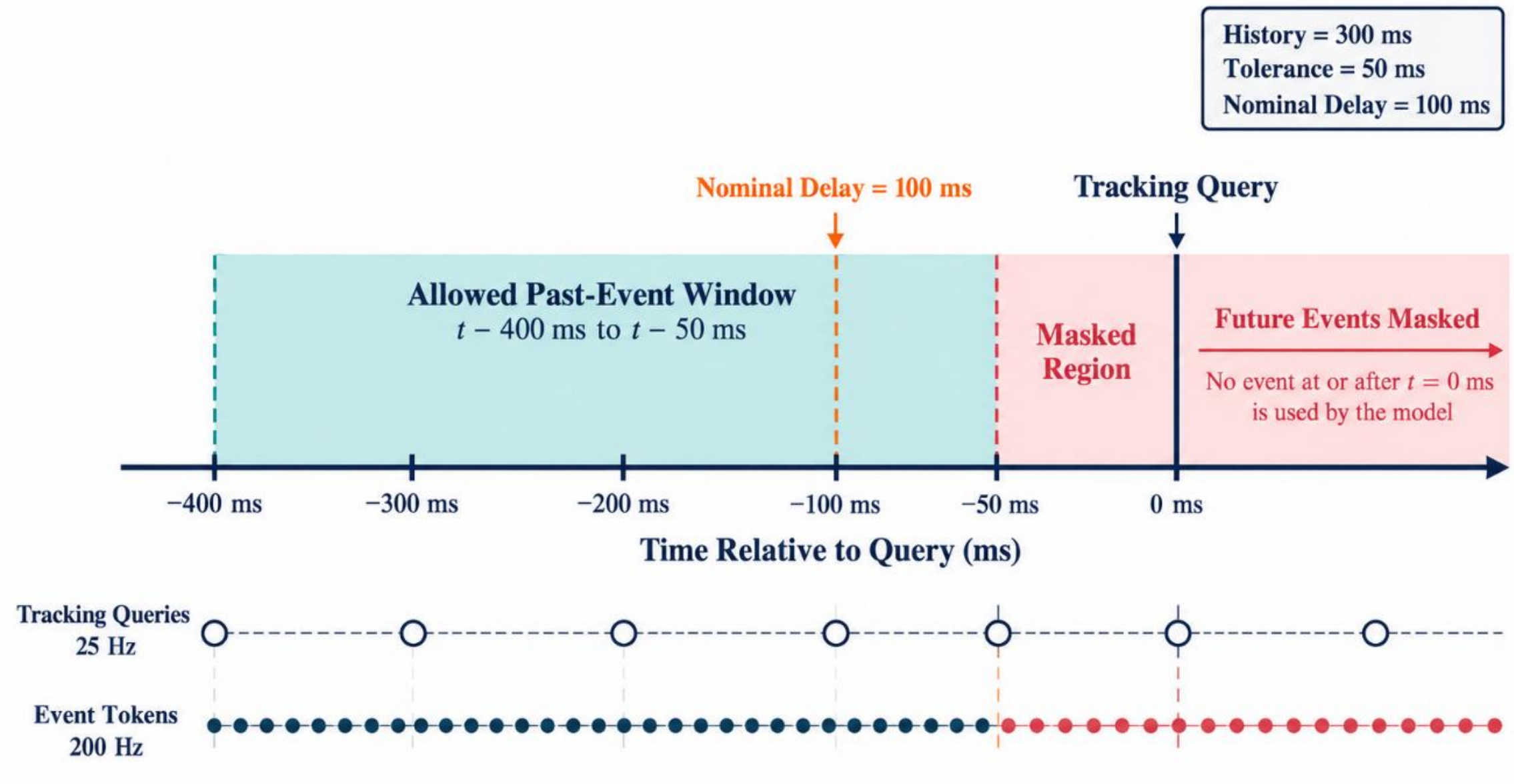

Figure 3. Temporal alignment: the valid event interval lies entirely before the query time.

Local attention learns weights within the permitted window rather than searching the full recording. A 100 ms nominal delay also does not restrict attention to a single event time. The tolerance forms a neighborhood, while the history parameter defines broader past context. These roles must be distinguished when interpreting delay sweeps: changing the nominal delay changes the events covered by the mask, while the event encoder, tracking features, and evaluation files remain fixed.

### 4.5 Causal Temporal Refinement and Rotary Position Encoding

Tokens produced by cross-attention pass through two causal RoPE encoder layers. Each layer has a model dimension of 128, four attention heads, a feed-forward width of 256, and dropout 0.10. RoPE encodes position by rotating query and key representations, allowing relative position to affect attention interactions [3]. The causal layers restrict each position to past and current information, while the cross-attention mask separately constrains the event-to-tracking temporal relationship.

The local cross-attention mask specifies the allowed lag between low-rate tracking queries and high-rate events. The causal self-attention mask limits access by refined tokens to information at other times. RoPE encodes relative positional information, while the masks keep temporal access consistent with the causal constraints.

### 4.6 Bounded Correction and Dynamic Gating

After temporal refinement, a one-dimensional convolution projects the token representation back to the 64 baseline feature channels. Its weights are initialized from a zero-mean normal distribution with standard deviation 0.001, and its bias is zero. The projected correction is then bounded by tanh. A second 1×1 convolution receives the concatenated baseline and correction features; its weights are initialized to zero and its bias to −2, after which a sigmoid produces a channel-wise gate. With zero initial weight contribution, the initial sigmoid value is sigmoid(−2) ≈ 0.119.

The output equals the baseline feature plus 0.15 times the element-wise product of the gate and bounded correction. The tanh function limits each correction component, while the dynamic gate controls correction strength across times and channels. Figure 4 shows this residual-fusion process.

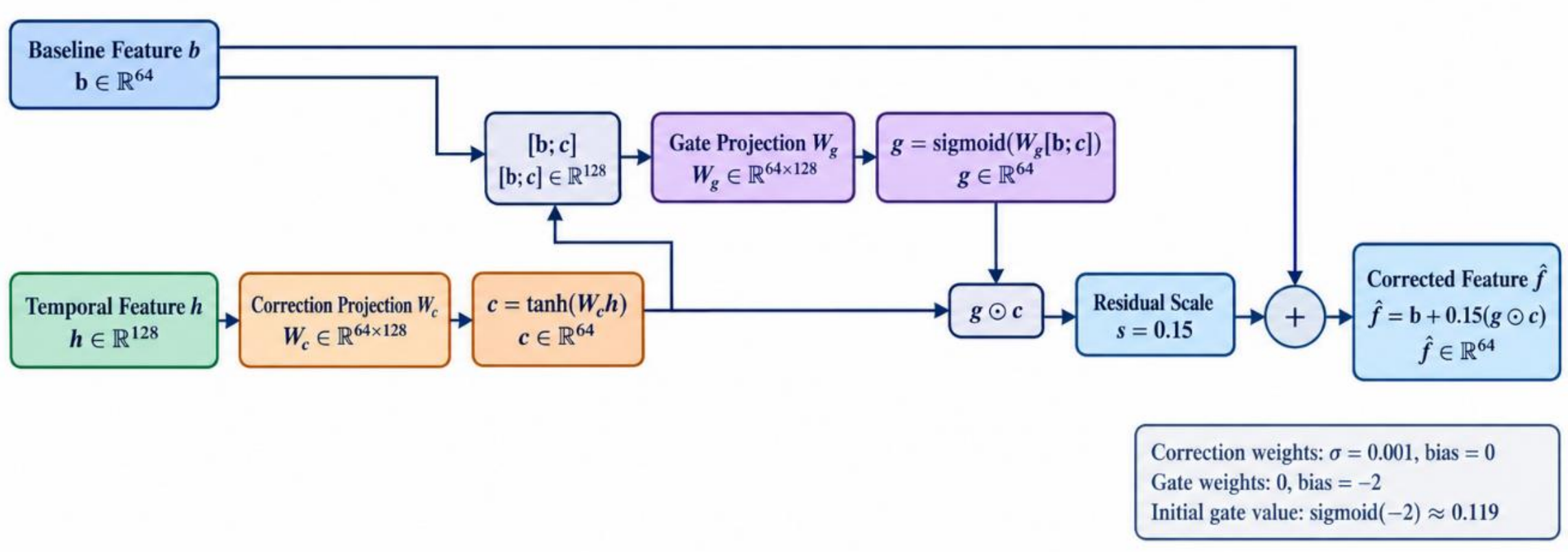

Figure 4. Bounded residual correction and dynamic gating.

### 4.7 Parameter and Timing Configuration

Table 1 summarizes the EFormer configuration. The system target is a 20-DoF hand pose, whereas feature output denotes the 64-channel representation returned by the feature network.

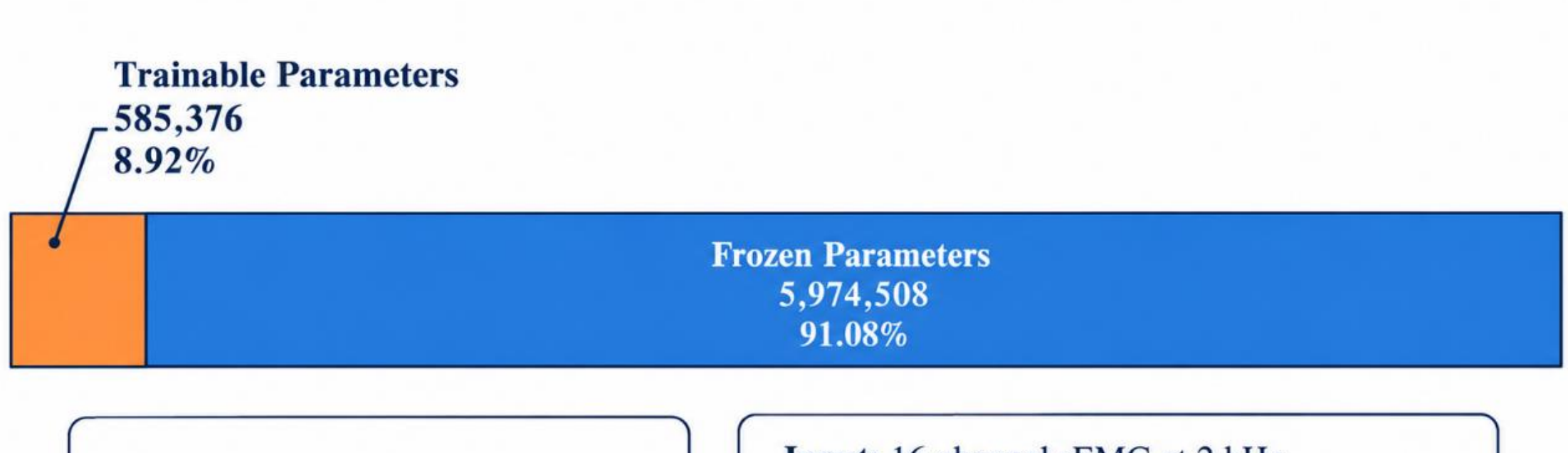

Figure 5. Architectural dimensions and trainable/frozen parameter counts for the evaluated configuration.

**Table 1. EFormer model configuration.**

| Component | Specification |
|---|---|
| Input | 16-channel sEMG, 2 kHz |
| System target | 20 hand-joint degrees of freedom |
| Feature output | 64 channels |
| Baseline feature rate | 25 Hz; stride 80 |
| Event input | raw signal + causal difference + 32-sample absolute envelope; 48 channels |
| Event encoding | 128 channels; stride 10; 200 Hz |
| Cross-attention | 128 dimensions, 4 heads, FFN 256, dropout 0.10 |
| Delay/history/tolerance | 100/300/50 ms |
| Effective event interval | 50–400 ms before the query; future events masked |
| Temporal encoder | 2 causal RoPE layers; 128 dimensions, 4 heads, FFN 256, dropout 0.10 |
| Correction | tanh projection; dynamic sigmoid gate; scale 0.15 |
| Continuation-training parameters | 585,376 trainable; 5,974,508 frozen |

## 5. Experimental Setup

### 5.1 Data and Metrics

The emg2pose benchmark provides 16-channel wrist sEMG sampled at 2 kHz with hand-pose labels, covering 193 users, 370 hours of recordings, and 29 movement stages [1]. EFormer is evaluated on continuous joint-angle prediction with 20 hand-joint degrees of freedom as the target. Test-set MAE, RMSE, and $R^2$ are reported.

Evaluation is based on errors between predicted joint angles and their corresponding labels. Test-set results compare EFormer with the official tracking baseline; validation MAE after five continuation-training epochs is reported separately as a training-process indicator.

### 5.2 Baseline and Main Comparison

The main comparison is between the official tracking baseline and EFormer with a fixed 100 ms nominal delay. Test performance for both models is summarized by joint-angle MAE, RMSE, and $R^2$. The validation result after five continuation-training epochs is reported separately.

## 6. Results

Table 2 reports test-set metrics for the official tracking baseline and EFormer.

**Table 2. Test results for the official tracking baseline and EFormer.**

| Setting | Evaluation split | MAE (rad) | RMSE (rad) | $R^2$ |
|---|---|---|---|---|
| Official tracking baseline | Test | **0.1745326** | **0.2715448** | **0.6791103** |
| EFormer, fixed 100 ms | Test | **0.1546634** | **0.24063** | **0.74801** |

Compared with the official tracking baseline, EFormer reduces test MAE from 0.1745326 rad to 0.1546634 rad, a relative reduction of 11.38%. RMSE decreases from 0.2715448 rad to 0.24063 rad, and $R^2$ increases from 0.6791103 to 0.74801. All three metrics move in the favorable direction, indicating lower overall error on this continuous pose-prediction task.

After five continuation-training epochs, the validation MAE is 0.1449901 rad. This value describes the training process and is not included in the test-set comparison in Table 2.

## 7. Discussion

### 7.1 Analysis of Performance

EFormer achieves a test-set MAE of 0.1546634 rad, an RMSE of 0.24063 rad, and an $R^2$ of 0.74801, outperforming the official tracking baseline on all three metrics. The relative MAE reduction is 11.38%. These results support the use of a locally time-aligned event branch for tracking-feature correction; Section 4 explains the structural role of each component.

### 7.2 Role of Local Temporal Alignment

At 2 kHz, the raw signal contains fine-grained temporal structure that is substantially compressed before the 25 Hz tracking representation is produced. The 200 Hz event path retains a denser summary of muscle activity. Cross-attention can then select the event tokens relevant to a low-rate tracking query from the permitted window. The architecture is therefore a form of multi-rate fusion that explicitly incorporates prior knowledge about physiological and annotation delays.

### 7.3 Frozen Baseline and Bounded Updates

Freezing the baseline keeps the initial tracking representation fixed while the correction branch is trained. The residual form allows the event branch to make incremental feature updates without replacing the baseline. The tanh projection bounds each correction component before gating, while the sigmoid gate controls its application strength. Zero-initialized gate weights and a negative bias reduce the initial contribution of the new path.

### 7.4 Relationship to the Benchmark

The published emg2pose study provides a large-scale benchmark covering held-out users, movement stages, and their combinations [1]. This work uses its continuous hand-pose prediction task and the official tracking system as the primary comparison. EFormer improves the reported joint-angle MAE, RMSE, and $R^2$.

REACT uses user-conditioned feature modulation to improve individual adaptation [10], whereas EFormer corrects tracking features through local temporal alignment and multi-rate event fusion. Because the two methods address different modeling factors, the official tracking baseline serves as the direct performance comparison.

## 8. Reproducibility Settings

### 8.1 Model Configuration

The EFormer feature network uses a frozen tracking backbone and a trainable correction branch. The continuation-training configuration contains 585,376 trainable parameters and 5,974,508 frozen parameters. Table 1 and Appendix A provide the principal network dimensions, temporal windows, and gating parameters.

### 8.2 Training Configuration

Continuation training uses five epochs, random seed 42, batch size 8, a learning rate of $1\times10^{-4}$, and a gradient-clipping threshold of 1.0. The input sampling rate is 2 kHz; the tracking and event strides are 80 and 10 samples, respectively; and the nominal delay, history, and tolerance are 100, 300, and 50 ms.

## 9. Conclusion

This paper presents EFormer, an sEMG feature-correction network built on a frozen tracking path. Its architecture includes a high-rate event encoder, locally time-aligned cross-attention, two causal RoPE temporal layers, and a dynamically gated bounded residual. The evaluated configuration uses a 100 ms nominal delay, and each query accesses only events from the preceding 50–400 ms.

On continuous hand-pose prediction, EFormer achieves an MAE of 0.1546634 rad, an RMSE of 0.24063 rad, and an $R^2$ of 0.74801, compared with 0.1745326 rad, 0.2715448 rad, and 0.6791103 for the official tracking baseline. EFormer reduces MAE by 11.38% relative to the baseline. These results show that residual fusion of a frozen tracking representation with temporally aligned event features improves system-level joint-angle prediction.

### Appendix A. EFormer Architecture and Hyperparameters

The EFormer tracking path uses 16 input channels and 64 output channels, with a left context of 1,790 samples, zero right context, and baseline dropout 0.01; the tracking backbone remains frozen. The event encoder outputs 128 event channels with dropout 0.05. Local attention receives 64-channel tracking features and 128-channel event features, projects both into a 128-dimensional model space, uses four attention heads, and applies dropout 0.10. The temporal stack contains two causal RoPE layers with a feed-forward width of 256 and dropout 0.10. Two 1×1 projections produce the correction and dynamic gate. The correction projection uses small-variance normal initialization with zero bias; the gate weights are initialized to zero and its bias to −2. The residual scale is 0.15.

This configuration specifies the principal EFormer hyperparameters and feature-computation path.

### Appendix B. Sampling and Time Conversion

Conversions between sample counts and temporal parameters provide a common time coordinate for event-to-tracking attention.

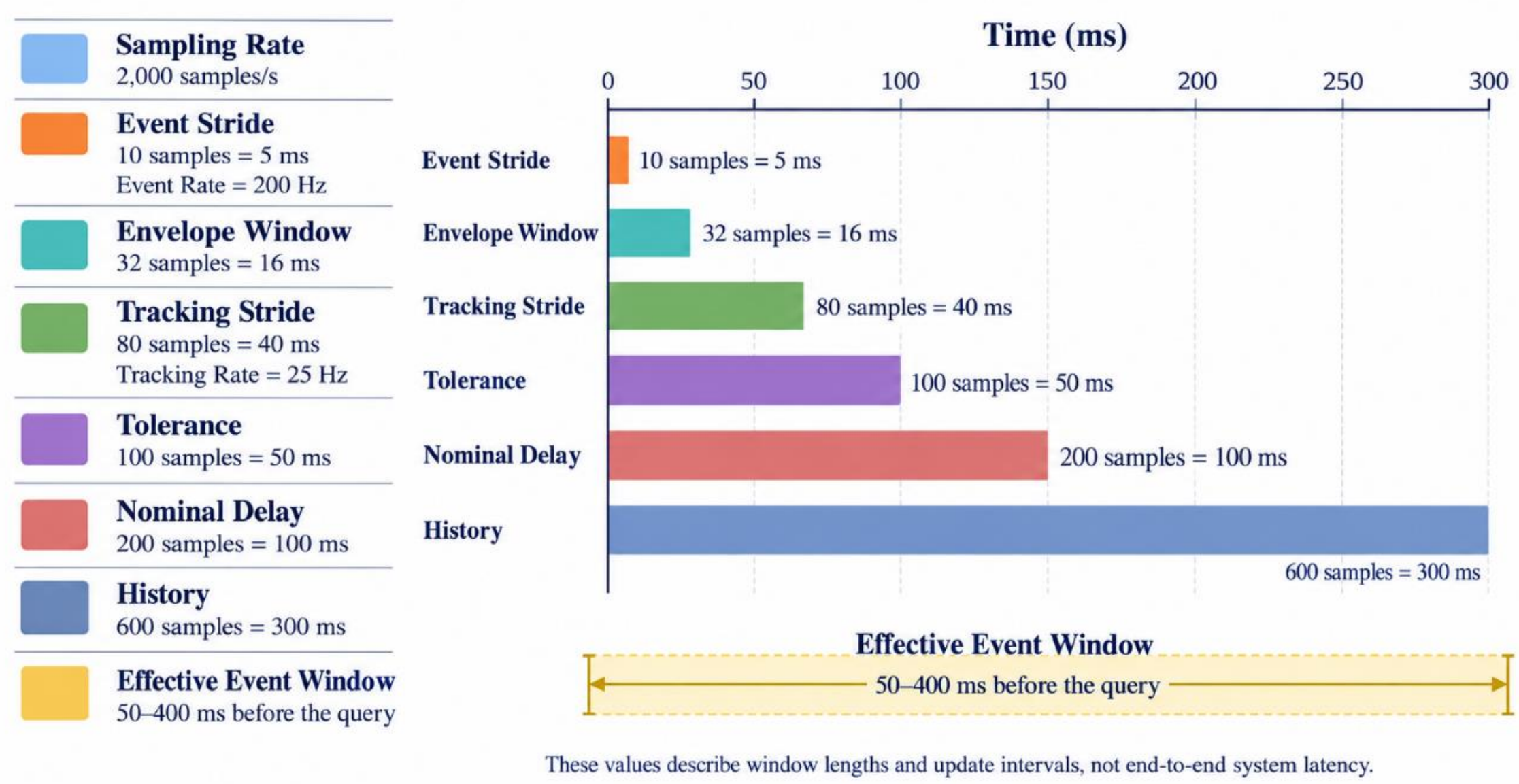

Figure 6. Sampling-rate and temporal-parameter conversion from raw samples to model steps.